\documentclass[runningheads]{llncs}
\usepackage[T1]{fontenc}
\usepackage{graphicx,verbatim}

\usepackage[utf8]{inputenc}
\usepackage[small]{caption}
\usepackage{graphicx}
\usepackage{amsmath}
\usepackage{booktabs}
\usepackage{algorithm}
\usepackage{algorithmic}
\usepackage[switch]{lineno}
\usepackage{amssymb} 
\usepackage{amsmath}
\usepackage{multirow}
\usepackage{colortbl}
\usepackage{soul}
\usepackage{pifont}
\usepackage{chngcntr}
\usepackage{float}
\usepackage{placeins}
\usepackage{afterpage}
\usepackage{url}
\usepackage{color}
\usepackage[colorlinks,bookmarksopen,bookmarksnumbered,citecolor=blue, linkcolor=blue, urlcolor=blue]{hyperref}
\usepackage{subcaption}  

\begin{document}
\title{Register Anything: Estimating ``Corresponding Prompts'' for Segment Anything Model}
%
\author{Shiqi Huang\inst{1,2} \and
Tingfa Xu\inst{2} \and
Wen Yan\inst{1} \and Dean Barratt\inst{1} \and Yipeng Hu\inst{1}}
\authorrunning{Shiqi Huang et al.}
%

\institute{University College London, London, UK \and Beijing Institute of Technology, Beijing, China
\\
\email{\{huangsq,ciom\_xtf1\}@bit.edu.cn}\\
\email{yipeng.hu@ucl.ac.uk}}

\maketitle              
\begin{abstract}
Establishing pixel/voxel-level or region-level correspondences is the core challenge in image registration. The latter, also known as region-based correspondence representation, leverages paired regions of interest (ROIs) to enable regional matching while preserving fine-grained capability at pixel/voxel level. Traditionally, this representation is implemented via two steps: segmenting ROIs in each image then matching them between the two images. In this paper, we simplify this into one step by directly ``searching for corresponding prompts'', using extensively pre-trained segmentation models (\textit{e.g.}, SAM) for a training-free registration approach, PromptReg. Firstly, we introduce the ``corresponding prompt problem'', which aims to identify a corresponding Prompt Y in Image Y for any given visual Prompt X in Image X, such that the two respectively prompt-conditioned segmentations are a pair of corresponding ROIs from the two images.  Secondly, we present an ``inverse prompt'' solution that generates primary and optionally auxiliary prompts, inverting Prompt X into the prompt space of Image Y.
Thirdly, we propose a novel registration algorithm that identifies multiple paired corresponding ROIs by marginalizing the inverted Prompt X across both prompt and spatial dimensions.
Comprehensive experiments are conducted on five applications of registering 3D prostate MR, 3D abdomen MR, 3D lung CT, 2D histopathology and, as a non-medical example, 2D aerial images. Based on metrics including Dice and target registration errors on anatomical structures, the proposed registration outperforms both intensity-based iterative algorithms and learning-based DDF-predicting networks, even yielding competitive performance with weakly-supervised approaches that require fully-segmented training data.

\keywords{Image registration  \and Corresponding representation \and Prompt engineering \and Segment Anything Model (SAM).}

\end{abstract}
\begin{figure*}[ht]
  \centering
  \includegraphics[width=0.85\textwidth]{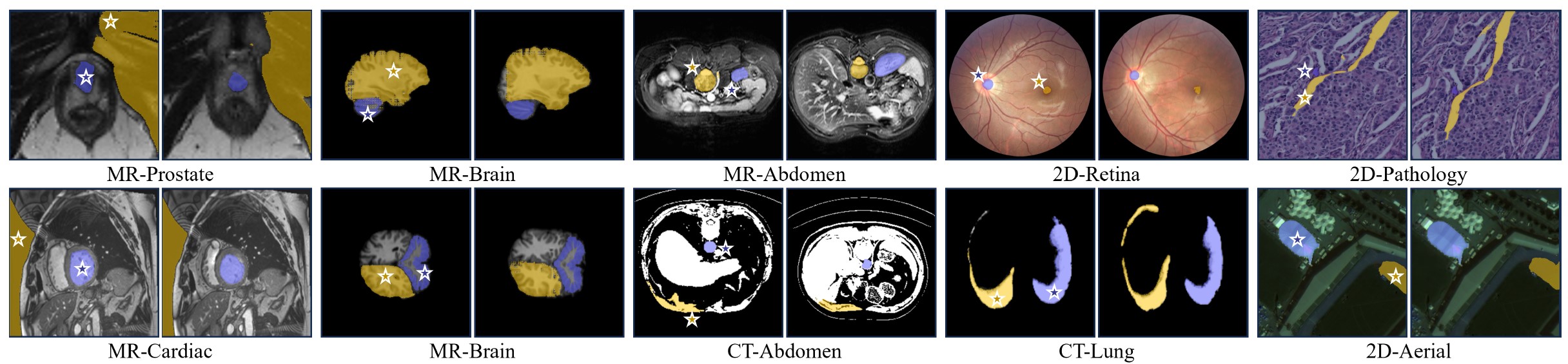}
  \caption{PromptReg performs generic registration tasks with ROI-based correspondence representation. Given any point prompt \ding{73}, it identifies corresponding ROIs in paired images, where the blue and yellow indicate prompt locations inside and outside anatomical ROIs, respectively.}
  \label{fig: illustration}
  \vspace{-2em}
\end{figure*}
\section{Introduction}


Image registration aligns two or more images to a common coordinate system, allowing for the overlay and analysis of corresponding features~\cite{de2019deep,eppenhof2018deformable,rohe2017svf}. This process is equivalent to establishing a ubiquitous correspondence at pixel/voxel level, that is, at each pixel/voxel location, a corresponding spatial location on another image is determined by registration. Pursing such ubiquitous correspondence under this premise, the correspondence is often formulated by parametric spatial transformation~\cite{rohr2001landmark}, such as rigid and deformable models, or dense samples of displacement vectors at all discrete pixel/voxel locations, \textit{i.e.}, dense displacement fields (DDFs)~\cite{detone2016deep,balakrishnan2019voxelmorph}.

Between medical images, specific meanings of the correspondence may vary, for example, to indicate the same anatomy scanned at different time points or structures that share homology from different subjects. As pointed out in previous works~\cite{hu2019conditional,huang2024one}, correspondence cannot (and need not) be defined necessarily everywhere (\textit{i.e.}, at every pixel/voxel) in many clinical applications. Consequently, a new ROI-based representation has recently been proposed. This correspondence representation led to a new registration algorithm, which utilised the Segment Anything Model (SAM) to segment multiple regions-of-interest (ROIs) in two images and matched them in feature space~\cite{huang2024one}, namely SAMReg. 

SAMReg relied on the SAM's unconditional segmentation capability to generate many candidates ROIs before matching their prototype features. However, this algorithm is inherently limited by its general-purposed formulation. First, the generated candidate ROIs cannot be controlled with respect to specific applications of interest. For instance, when prior knowledge on which class of ROIs should be considered more important, there is no mechanism to incorporate it. Second, the prototype matching is based on similarity (\textit{e.g.} cosine similarity) between features of the candidate ROIs. This may be effective for a specific application, using carefully tuned hyperparameters such as the prototype dimensionality and the similarity threshold, but does not guarantee that the matched ROIs are of the same class between the two images.

In this study, we first develop methodologies for searching corresponding ROI pairs of the same class. Unlike the previously proposed unconditioned algorithm~\cite{huang2024one}, the proposed approach exploits the prompt-conditioned SAM for, arguably, more controllable ROI class definitions. These classes may be i) predefined by prior clinical requirements to aid application-specific registration algorithms or facilitate an interactive registration; but can also be ii) randomly sampled for general-purpose, no-prior image registration algorithms. Focusing on point-based prompts, we formalize this as a ``corresponding prompt problem'' in Sec.~\ref{sec: problem_def}. We then propose a solution by inverting a given prompt in Sec.~\ref{sec: prompt_setting}. 

With prompt-based segmentation, the inherent limitation in SAM and its generalisation to medical images became evident. For example, sensitivity in prompt localisation leads to unstable segmented ROI was reported~\cite{he2024weakly,ke2024segment}, which was found more predominant in a domain-shifted dataset, such as medical images, and needed to resort a nontrivial fine-tuning effort~\cite{ma2024segment,wu2023medical}. For our registration algorithms, we further propose a marginalization strategy, over both the prompt sampling space and the transformation space, in Sec.~\ref{sec: marginalisation}.

\begin{figure*}[!t]
  \centering
  \includegraphics[width=0.8\textwidth]{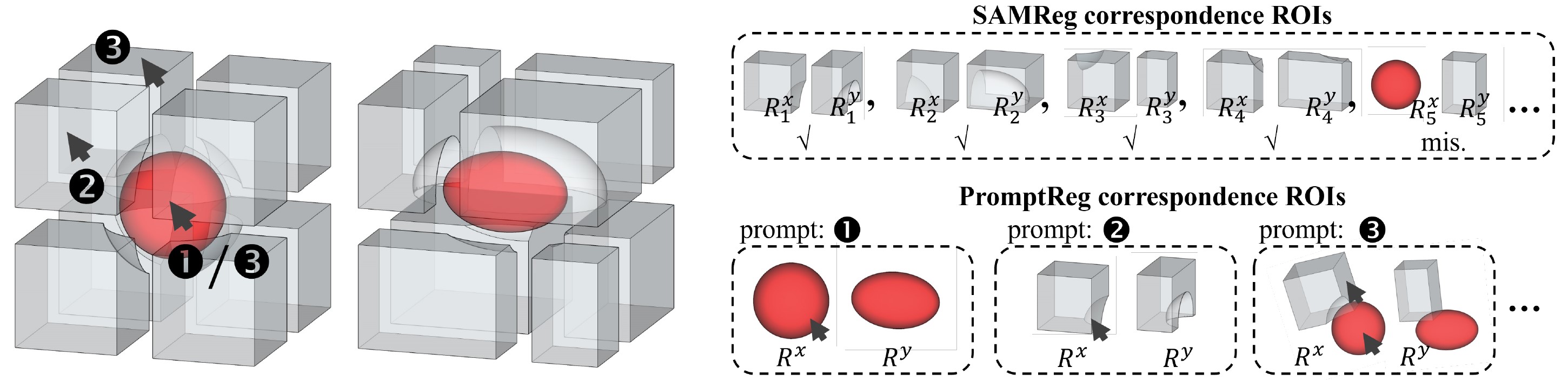}
  \caption{Comparison of correspondence ROIs from different paradigms, where the red denotes the targeted ROI and the arrow represents the given prompts.}
  \label{fig:correspondence}
  \vspace{-2em}
\end{figure*}

In summary, our main contributions are: 
\textbf{1)}We propose a new image registration paradigm, which seeks correspondence-conditioned ROI pairs from two respective images. To our knowledge, it is the first promptable registration trial.
\textbf{2)} Using SAM without any adaptation or fine-tuning, we introduce the general-purpose PromptReg by segmenting corresponding ROIs from a pair of images, based on any (manual or randomly-sampled) given vision prompts.
\textbf{3)} Extensive experiments show that the proposed registration algorithms outperform the commonly-adopted iterative registration and unsupervised registration, and is competitive with weakly-supervised registration 
across five clinical applications.

\section{Corresponding Prompt Problem}
\label{sec: problem_def}
\textbf{Revisiting ROI-based Correspondence Representation.}
The ROI-based correspondence representation for registration~\cite{huang2024one}, spatial locations $\textbf{X}$ and $\textbf{Y}$ within a moving and a fixed image, $I^x$ and $I^y$, respectively, is considered corresponding ROIs, represented through $K$ pairs of ROI $\{(R^{x}_k, R^{y}_k)\}_{k=1}^{K}$. Here, $R^{x}_k = \{\textbf{x}_l\}_{l=1}^{L^x_k}$ and $R^{y}_k = \{\textbf{y}_l\}_{l=1}^{L^y_k}$ are two sets of $L^x_k$ and $L^y_k$ spatial discrete locations (\textit{e.g.}, pixels/voxels) in the respective moving and fixed image spaces, $\textbf{x}_l\in\textbf{X}$ and $\textbf{y}_l \in \textbf{Y}$. This representation is thought necessary and effective for clinical practice. However, a dense correspondence representation, such as the DDF $\mathcal{T}$, can also be obtained by refining the region-specific alignment measure $\sum_{k=1}^K\mathcal{L}_{roi}(R^y_k, \mathcal{T}(R^x_k,\Theta))$, where $\Theta$ is parameters of the transformation $\mathcal{T}$.

The corresponding ROIs can be acquired through two ways: (1) \textit{segmenting known shared classes}, or (2) \textit{segmenting unknown classes and then matching them}. Currently, the former is conducted by weakly-supervised registration methods~\cite{hu2018weakly,hu2019conditional}, which train the segmentor with predefined $K$-class groundtruth to predict intuitive corresponding ROIs of the same classes $\textbf{C}=[C_1,...,C_K]^\top$, $\textbf{C}= \textbf{C}^x=\textbf{C}^y$, albeit with limited number of representation ROIs and narrow generalizability. 
The latter paradigm is implemented by recent SAMReg~\cite{huang2024one}, consisting two independent segmentation without conditioning prompts, producing finite $K^x$ and $K^y$ classes $\textbf{C}^x$ and $\textbf{C}^y$, with the highest class probabilities, respectively. Second, the subsequent \textit{matching} identifies $K$ shared classes $\textbf{C}$, and $\{C_k\}_{k=1}^K = \{C^x_k\}_{k=1}^{K^x} \cap \{C^y_k\}_{k=1}^{K^y}$. In practice, it is this two-step approximation that limits the SAMReg to represent a required ROI correspondence that are locally accurate and densely sampled. 

\noindent\textbf{Why ``Searching Corresponding Prompts'' Instead of ``Matching Corresponding ROIs''?}
A mirrored paradigm of SAMReg for \textit{segmenting known shared classes at the same time} is to estimate two vision prompts $Z^x_k$ and $Z^y_k$, for $I^x$ and $I^y$, respectively, such that the common class $C_k$ can be segmented by conditioning two SAM inferences using these two, namely, corresponding prompts. Compared with predefined classes in weakly-supervised methods, the class $C_k$, conditioned on prompt $Z^x_k$, requires neither specific labels nor anatomical knowledge, while still permitting targeted human interaction via $Z^x_k$.



The ``Searching Corresponding Prompts'' pattern is more cost-effective than SAMReg’s ``Matching Corresponding ROIs,'' both globally and locally. SAMReg cannot ensure that segmented classes from two independent SAM inferences intersect, $\{C^x_k\}_{k=1}^{K^x} \cap \{C^y_k\}_{k=1}^{K^y} \neq \varnothing,$ nor guarantee a specific local ROI class $C^*$ exists in both, $C^* \in (\{C^x_k\}_{k=1}^{K^x} \cap \{C^y_k\}_{k=1}^{K^y}).$ In contrast, corresponding prompts resolve empty intersections by leveraging predefined (manual or automatic) target classes $C^*$ (Fig.~\ref{fig:correspondence}).  
For dense correspondence, SAMReg lacks control over ROI sampling. Even with many candidate ROIs, there is no guarantee that they or the matched set $\textbf{C}$ will align with the targeted ROIs. The finite sizes of $\textbf{C}^x$ and $\textbf{C}^y$ further constrain $\textbf{C}$, increasing mismatches as more are assigned to $\textbf{C}$.

\noindent\textbf{What is the Definition of ``Corresponding Prompt Problem''?}
Given images $I^x$ and $I^y$, the goal is to estimate same-class ROIs. A vision prompt for $I^x$ is $Z^x_k = \{\textbf{x}_p\}_{p=1}^{P_k^x}$, where $\textbf{x}_p \in \textbf{X}$ and $P_k^x$ is the number of discrete pixels/voxels. The pre-trained SAM segments ROI as $R^x_k = f^{SAM}(I^x, Z^x_k),$ assuming $R^x_k$ belongs to class $C_k$.  

The aim is to search for the ``corresponding prompt'' $Z^y_k = \{\textbf{y}_p\}_{p=1}^{P_k^y}$, defined by $P_k^x$ locations $\textbf{y}_p \in \textbf{Y}$ in the second image $I^y$, such that the segmented ROI $R^{y}_k= f^{SAM}(I^{y},Z^y_k)$ in image $I^y$ mirrors the same class $C_k$ of the ROI $R^{x}_k$ in image $I^x$ and $(R^{x}_k,R^{y}_k)$ is considered as the ``corresponding ROI''.
This paradigm allows for the investigation of class-consistency across different images, leveraging the capabilities of the SAM or its variants to perform corresponding ROI segmentation, based on either predefined or arbitrary visual prompts.

\section{PromptReg: Promptable Registration Algorithm via Inverse Prompt Engineering}

The proposed PromptReg comprises two steps: (inverted) prompt searching and prompt marginalization. In the first step, prompt searching infers the corresponding prompt on the target image, while in the second step, prompt marginalization leverages an augmentation-based strategy to improve robustness.

\subsection{Prompt Searching}
\label{sec: prompt_setting}
Prompt searching aims to map a given prompt $Z_k^x$ from an image $I^x$ to a corresponding prompt set $Z^y_k$ on the other image $I^y$, ensuring both represent the same class $C_k$. In this section, the prompt set $Z^y_k$ is set to include a primary prompt, derived through an inversion process linking $C_k$ to $I^y$, and an optional auxiliary prompt. 
When the primary prompt alone cannot effectively represent $C_k$, (e.g., due to large image differences or low prominence of $C_k$), the auxiliary prompt, located at the foreground-background boundary, supplements $Z^y_k$.

\noindent\textbf{Primary Prompt.}
In a typical SAM inference for segmenting ROI \( R^x_k \) from image \( I^x \), an image encoder \( \mathcal{E}^{im} \) embeds \( I^x \in \mathbb{R}^{H\times W\times D} \) into features \( F^x \in \mathbb{R}^{H^\prime \times W^\prime \times D^\prime\times N^\prime} \). A prompt encoder \( \mathcal{E}^{pr} \) embeds the prompt \( Z^x_k \) into a class encoding \( C_k \), which may be unknown or unspecified, determined by \( Z^x_k \). The mask decoder \( \mathcal{D} \) then decodes \( F^x \), conditioned on \( C_k \), to produce the segmented ROI \( R^x_k \).
$R^x_k$ is further abstracted as a prototype $G_k = f^{proto.}(R^x_k,F^x)$, where $f^{proto.}$ is the prototype extraction function iterating all voxel locations, commonly used in prototypical few-shot learning~\cite{dong2018few,huang2023rethinking,chen2025augmenting,yang2018robust}. The resulting prototype $G_k\in\mathbb{R}^{1\times N^\prime}$ serves as an alternative representation of the class $C_k$.

$G_k$ is considered a function of only one variable $Z^y_k$, \textit{i.e.},
    $G_k=f(Z^x_k)=f^{proto.}(\mathcal{D}(F^x,\mathcal{E}^{pr}(Z^y_k)),F^x).$
Using the first-order Taylor expansion $G_k \approx f^0 + \mathcal{J} \cdot Z^y_k$ obtains a general solution for $Z^y_k \propto \mathcal{J}^{-1} \cdot G_k$, where $f^0$ is a constant and $\mathcal{J}\in \mathbb{R}^{N^\prime \times 3}$ (given 1 point prompt) is the Jacobian with respect to the function $f$, specifically, $\mathcal{J} =  \frac{\partial f}{ \partial Z^x_k}
        =  \frac{\partial f^{proto.}}{\partial \mathcal{D}} \cdot \frac{\partial \mathcal{D}}{\partial \mathcal{E}^{pr}} \cdot \frac{\partial \mathcal{E}^{pr}}{\partial Z^x_k}.$

To inverse the inference with image $I^y$ and unknown prompt $Z^y_k$, the logits of class probability maps $S^{\prime{x}}_k,  S^{\prime{y}}_k,\in \mathbb{R}^{H^\prime \times W^\prime \times D^\prime}$ are generated using cosine similarity function $f^{sim}$, \textit{e.g.}, $S^{\prime{x}}_k = f^{sim}(G_k, F^x), \text{where} ~S^{\prime{x}}_{k{(\textbf{i}^\prime)}} = \frac{G_k\cdot F^x_{(\textbf{i}^\prime)}} {\lVert G_k\rVert\cdot\lVert F^x_{(\textbf{i}^\prime)}\rVert}, $
where $\textbf{i}^\prime \in \mathbb{R}^3$ are all the 3D ``spatial'' locations in the feature space. The bracket subscripts $(\textbf{i}^\prime)$ is denoted to index the scalars and vectors, from $S^{\prime{x}}_k$ and $F^x$, respectively. With the same $G_k$, $S^{\prime{y}}_k = f^{sim}(G_k, F^y)$ is computed. 

Following the Taylor expansion, $S^{\prime{y}}_k \approx (f^{sim})^0 + \mathcal{J}^{sim}\cdot G_k$, 
leads to the association between $S^{\prime{y}}_k$ and $Z^y_k$ that
\begin{equation}
    Z^y_k = (\mathcal{J}^{sim})^{-1}\cdot \mathcal{J}^{-1} \cdot S^{\prime{y}}_k+\rho,
\end{equation}
where $\rho$ is a constant set manually and the similarity Jacobian matrix $\mathcal{J}^{sim}\in \mathbb{R}^{H^\prime W^\prime D^\prime \times N^\prime}$ is derived by $\mathcal{J}^{sim} = \frac{\partial S^{\prime{y}}_k}{\partial G_k}, \text{where}~
    \frac{\partial S^{\prime{y}}_{k(\textbf{i}^\prime)}}{\partial G_k} = \frac{F^x_{(\textbf{i}^\prime)}-S^{\prime{x}}_{k{(\textbf{i}^\prime)}}\cdot\frac{G_k}{\lVert G_k\rVert}}{\lVert F^x_{(\textbf{i}^\prime)}\rVert\cdot\lVert G_k\rVert}.$

The obtained inverse prompt $Z^y_k$ corresponds to the given prompt $Z^x_k$ with the same structural format, \textit{i.e.}, it adheres to a one-to-one or multiple-to-multiple correspondence between points.

\noindent\textbf{Auxiliary Prompts.}
Conditioned on the primary prompt $Z^y_k$, the segmented ROI $R^y_k=\mathcal{D}(F^y,\mathcal{E}^{pr}(Z^y_k)$ is generated. Ideally, $R^y_k$ should closely match the corresponding ROI, $R^x_k$, produced from prompt $Z^x_y$. However, $R^x_k$ and $R^y_k$ may not align in some cases, 
for example, $R^y_k$ 
partially capture a complete structure identified by $R^x_k$. To address these mismatches, more precise prompts are necessary to define the desired ROI $R^y_k$.

Given the inherent consistency of registration tasks, misaligned ROIs ($R^x_k$ and $R^y_k$) exhibit abnormal shapes. To quantify this, we compute the Hausdorff distance $d_H^{x \rightarrow y}$ and $d_H^{y \rightarrow x}$ between their contours, identifying points of maximum discrepancy $\textbf{x}^*$ and $\textbf{y}^*$. If $\|d_H^{x \rightarrow y}-d_H^{y \rightarrow x}\| > \sigma$, where $\sigma$ is empirically determined, $\textbf{x}^* ~\text{or}~ \textbf{y}^*$ is considered an anomaly. 

The prompt is iteratively placed until convergence, \textit{i.e.}, $\|d_H^{x \rightarrow y}-d_H^{y \rightarrow x}\| < \sigma$. In $s$-th iteration, based on the identified anomaly, we generate a new prompt, $\textbf{z}^y_{k,s}$. This prompt can be either a negative point or a positive point:
\begin{equation}
\label{eq:aux_prompt}
    \textbf{z}^y_{k,s} = \begin{cases} 
\mathbf{y}^* ~\text{(neg)}, & d_H^{x \rightarrow y} > d_H^{y \rightarrow x}, \\
\mathbf{x}^* + \epsilon \nabla f^{x \rightarrow y}(\mathbf{r}) \big|_{\mathbf{r} = \mathbf{x}^*} ~\text{(pos)}, & d_H^{x \rightarrow y} < d_H^{y \rightarrow x},
\end{cases}
\end{equation}
where $f^{x\rightarrow y}(\textbf{r})=\frac{1}{2}\|\textbf{r}-\textbf{y}^*\|^2$. These prompts are iteratively incorporated into $Z^y_k$ to progressively align $R^y_k$ with the $R^x_k$.


\subsection{Prompt Marginalization}
\label{sec: marginalisation}

\textbf{Probabilistic Notes.} The SAM models class probability as $\mathcal{R}_k = p( C_k | \textbf{I}, \textbf{Z}_k).$ Marginalizing over $\textbf{Z}_k$ and $p(C_k | \textbf{I}) = \int_{\textbf{Z}_k \in \Omega_{Z^x_k}} p(C_k | \textbf{I}, \textbf{Z}_k) p(\textbf{Z}_k | \textbf{I}) d\textbf{Z}_k,$ where $\Omega_{Z^x_k}$ defines the prompt sampling space on $I^x$. Multiple prompts enable marginalization, capturing $\textbf{Z}_k$ uncertainty and improving spatial coverage (\textit{i.e.}, $K>1$).  

For a second image $I^y$, estimating $\textbf{Z}_k=Z^y_k$ considers it a parameter linked to the prototype $G^x_k$ from $Z^x_k$ (Sec.~\ref{sec: prompt_setting}). The marginal likelihood with fixed $I^y$ is: $\int_{\textbf{R}^y_k \in \Omega_{R^y_k}} p(\textbf{R}^y_k | Z^y_k, I^y) d\textbf{R}^y_k = \int_{\textbf{G}^x_k \in \Omega_{G^x_k}} p(\textbf{G}^x_k | Z^y_k, I^y) d\textbf{G}^x_k= \int_{{\textbf{Z}_k \in \Omega_{Z^x_k}}, \textbf{I} \in \Omega_{I^x}} p\\(\textbf{Z}_k, \textbf{I} | Z^y_k, I^y) d(\textbf{Z}_k, \textbf{I}). $ 
Here, $\textbf{R}^y_k$ and $\textbf{G}^x_k$ denote the random variables of segmented ROI and prototype, and $\Omega_{*}$ defines the sampling space. Estimating this marginal likelihood follows Bayesian model averaging~\cite{raftery1997bayesian}.

\noindent\textbf{Spatial Transformation for Aggregation.} 
To sample $\textbf{Z}_k$ and $\textbf{I}$ from respective $\Omega_{Z^x_k}$ and $\Omega_{I^x}$, one could argue the benifit in applying independent spatial transformation $\mathcal{A}_j$ to each. However, to ensure the transformed prompt $\tilde{Z}^x_{k,j}=Z^x_k \circ \mathcal{A}_j$ and image $\tilde{I}^x_j=I^x \circ \mathcal{A}_j$ represent the same class $C_k$, the same spatial transformation $\mathcal{A}_j$ is applied.

Based on the $J$ transformed images $\{\tilde{I}^x_j\}$ and prompts $\{\tilde{Z}^x_{k,j}\}$, a set of ROIs (class probability maps) $\{\tilde{\mathcal{R}}^x_{k,j}\}$ are predicted. Using the inverse-transformed ROIs $\{\tilde{\mathcal{R}}^x_{k,j} \circ \mathcal{A}_j^{-1}\}$, a set of corresponding prompts are computed $\{\tilde{Z}^y_{k,j}\}$, which obtains a set of corresponding ROIs $\{\tilde{\mathcal{R}}^y_{k,j}\}$, for each class $C_k$. The final pair of corresponding ROIs $(\bar{\mathcal{R}}^x_k, \bar{\mathcal{R}}^y_k)$ are averages:  $\bar{\mathcal{R}}^x_k = \frac{1}{J}\sum_{j=1}^J \tilde{\mathcal{R}}^x_{k,j},     \bar{\mathcal{R}}^y_k = \frac{1}{J}\sum_{j=1}^I \tilde{\mathcal{R}}^y_{k,j}.$



\section{Experiments and Results}
\textbf{Datasets and Evaluation Metrics:}
We evaluate our method in five medical and non-medical datasets: MR-Prostate~\cite{ahmed2017diagnostic}, MR-Abdomen~\cite{kavur2021chaos}, CT-Lung~\cite{LUNG}, 2D-Pathology~\cite{HISTOLOGY} and 2D-Aerial~\cite{AERIAL}.
The first three datasets are three-dimensional and contain MR and CT imaging modalities, while the latter two are two-dimensional.
To ensure a fair comparison with other supervised registration methods, 80\% of the images are utilized for training purposes. Our method does not use this subset for training and is tested on the remaining data alongside other methods. Additionally, for ablative experiments, the proposed algorithm is implemented to all available images. 
Regarding registration strategies, inter-subject registration is applied to the MR-Prostate and MR-Abdomen datasets to align images to a standard reference. Conversely, intra-subject registration applies for the lung, histological, and aerial images to align image pairs that acquired at varying time.
Metrics include Dice and target registration error (TRE). 
For non-rigid registrations, Dice is assessed on moved and fixed critical anatomical structures, with TRE derived from the centroids of these ROIs.

\noindent\textbf{Implementation Details:} 
In our evaluation experiments, the prompt is randomly sampled four times to infer dense correspondence. Hyperparameters $\sigma$ and $\epsilon$ for auxiliary prompt setting is set to $20.0$ and $2.0$, respectively. For marginalization in spatial transformation, transformations such as flipping, rotation, and scaling are randomly sampled, with flipping consisting of horizontal and vertical, rotation ranging from $0^{\circ}$ to $360^{\circ}$, scaling ranging from $0.5$ to $2.5$. The implementation is based on PyTorch and MONAI~\cite{cardoso2022monai}. The algorithm was executed on an NVIDIA Quadro GV100. A code demo is available at: 
\href{https://github.com/sqhuang0103/PromptReg.git}{PromptReg}.

\begin{table}
\centering
\scriptsize
\setlength{\tabcolsep}{2pt}
\renewcommand{\arraystretch}{1.2} 
\resizebox{\textwidth}{!}{%
\begin{tabular}{c|cc|cc|cc|cc|cc} 
\hline
\multirow{2}{*}{\textbf{Methods }} & \multicolumn{2}{c|}{\textbf{MR-Prostate }} & \multicolumn{2}{c|}{\textbf{MR-Abdomen }} & \multicolumn{2}{c|}{\textbf{CT-Lung }} & \multicolumn{2}{c|}{\textbf{2D-Pathology }} & \multicolumn{2}{c}{\textbf{2D-Aerial }} \\ 
\cline{2-11}
 & \textbf{Dice} & \textbf{TRE} & \textbf{Dice} & \textbf{TRE} & \textbf{Dice} & \textbf{TRE} & \textbf{Dice} & \textbf{TRE} & \textbf{Dice} & \textbf{TRE} \\ 
\hline
\textbf{NiftyReg}\cite{modat2014global} & 7.68±3.98 & 4.67±3.48 & 8.93±2.21 & 3.13±2.89 & 10.93±2.02 & 4.23±1.64 & 6.81±3.02 & 5.90±3.75 & 10.21±3.02 & 4.02±2.25 \\
\rowcolor[rgb]{0.753,0.753,0.753} \textbf{VoxelMorph}\cite{balakrishnan2019voxelmorph} & 55.94±3.34 & 3.68±1.98 & 58.1±3.95 & 2.76±2.41 & 77.98±2.72 & 3.24±0.81 & 59.34±3.72 & 4.31±2.13 & 72.73±2.41 & 3.53±1.30 \\
\textbf{LabelReg}\cite{hu2018weakly} & 76.72±3.23 & 2.72±1.23 & 75.97±2.42 & 1.56±1.34 & 83.56±2.43 & 1.52±0.86 & - & - & - & - \\
\rowcolor[rgb]{0.753,0.753,0.753} 
\textbf{KeyMorph}\cite{evan2022keymorph} & 70.52±3.25 & 3.19±1.63 & 72.77±2.56 & 2.21±1.87 & 81.71±2.26 & 2.93±0.86 & 73.15±2.39 & 3.47±2.01 & 80.22±2.13 & 3.05±1.19 \\
\textbf{TransMorph}\cite{chen2022transmorph} & 71.77±3.01 & 3.11±1.27 & 73.81±2.52 & 2.10±1.76 & 81.97±2.19 & 2.84±0.79 & 75.42±2.08 & 3.29±1.95 & 82.63±2.09 & 2.91±1.02 \\
\rowcolor[rgb]{0.753,0.753,0.753} \textbf{SAMReg}\cite{huang2024one} & 75.67±3.19 & 2.09±1.22 & 73.65±2.52 & 1.43±1.21 & 85.23±2.16 & 1.31±0.91 & 69.87±3.70 & 3.12±1.23 & 85.34±2.31 & \ul{2.57±1.10} \\ 
\hline
\multicolumn{1}{l|}{\textbf{PromptReg} \textit{w} \textbf{SAM}\cite{kirillov2023segment}} & \ul{76.48±2.92} & \ul{2.10±2.68} & 72.48±3.57 & 2.38±1.13 & 87.76±3.21 & 2.02±1.03 & 72.71±3.63 & 3.41±0.91 & \ul{86.65±1.80} & 2.62±1.01 \\
\rowcolor[rgb]{0.753,0.753,0.753} \multicolumn{1}{l|}{\textbf{PromptReg} \textit{w} \textbf{MedSAM}\cite{ma2024segment}} & 72.91±3.12 & 3.32±2.30 & \ul{73.86±3.37} & 2.18±1.22 & \ul{88.40±3.23} & \ul{1.66±0.99} & 71.47±3.52 & \ul{3.14±1.34} & 83.77±2.03 & 3.69±1.78 \\
\multicolumn{1}{l|}{\textbf{PromptReg} \textit{w} \textbf{SAMed2d}\cite{cheng2023sammed2d}} & 74.28±3.39 & 3.70±2.04 & 74.55±3.54 & \ul{1.86±1.05} & 88.03±3.10 & 2.06±0.84 & \ul{87.35±3.89} & 3.90±1.97 & 82.69±2.13 & 2.86±1.65 \\ 
\hline
\rowcolor[rgb]{0.753,0.753,0.753} \multicolumn{1}{l|}{\textbf{PromptReg} \textit{w} \textbf{SAMed3d}\cite{wang2023sammed3d}} & 75.63±2.61 & 3.95±1.59 & 75.35±2.99 & 1.72±1.01 & 88.44±2.83 & 1.73±0.84 & - & - & - & - \\
\multicolumn{1}{l|}{\textbf{PromptReg} \textit{w} \textbf{3dAdapter}\cite{wu2023medical}} & \textbf{77.76±2.20} & \textbf{2.06±0.91} & \textbf{77.3±2.53} & \textbf{1.22±0.92} & \textbf{90.46±2.53} & \textbf{1.20±0.82} & - & - & - & - \\
\hline
\end{tabular}
}
\caption{Comparison of different registration models and our PromptReg with various promptable models across five datasets, highlighting the best 3D model results in \textbf{bold} and the best 2D model results with an \ul{underline}.}
\vspace{-2em}
\label{tab:sota}
\end{table}

\begin{table}[t]
\centering
\scriptsize
\setlength{\tabcolsep}{2pt}  
\renewcommand{\arraystretch}{1.2}  

\begin{minipage}{0.45\textwidth}  
    \centering
    \resizebox{\textwidth}{!}{
    \begin{tabular}{cc|cc} 
    \hline
    \multicolumn{2}{c|}{\textbf{Prompt Location}} & \multirow{2}{*}{\textbf{Dice}} & \multirow{2}{*}{\textbf{TRE}} \\ 
    \cline{1-2}
    \textbf{outsideROI} & \textbf{insideROI} &  &  \\ 
    \hline
    \checkmark & \checkmark & 77.76±2.20 & 2.06±0.91 \\
    \rowcolor[rgb]{0.753,0.753,0.753} \checkmark &  & 75.23±3.11 & 2.83±1.06 \\
     & \checkmark & 81.77±3.08 & 1.90±0.91 \\
    \hline
    \end{tabular}}
    \caption{Ablation of given prompt $Z^x_k$ location on MR-Prostate dataset, where \textit{ROI} denotes the target anatomical region.}
    \label{tab:x_loc}
\end{minipage}
\hfill
\begin{minipage}{0.45\textwidth}  
    \centering
    \resizebox{\textwidth}{!}{
    \begin{tabular}{c|cc} 
    \hline
    \textbf{Aux Threshold $\sigma $}(pix) & \textbf{Dice} & \textbf{TRE} \\ 
    \hline
     10.0 & 75.12±2.32 & 2.29±1.01 \\
    \rowcolor[rgb]{0.753,0.753,0.753}20.0 & 77.76±2.20 & 2.06±0.91 \\
     30.0 & 76.43±2.29 & 2.07±0.97 \\
    \rowcolor[rgb]{0.753,0.753,0.753}40.0 & 74.79±2.43 & 2.11±1.02 \\
     50.0 & 72.50±2.51 & 2.18±1.08 \\
    \rowcolor[rgb]{0.753,0.753,0.753}$+\infty$ (pri. only) & 70.12±2.96 & 3.12±1.31 \\
    \hline
    \end{tabular}}
    \caption{Ablation of threshold $\sigma$ for inverse prompt $Z^y_k$ on MR-Prostate dataset.}
    \label{tab:y_form}
\end{minipage}
\\
\begin{minipage}{0.35\textwidth}  
    \centering
    \resizebox{\textwidth}{!}{
    \begin{tabular}{c|cc} 
    \hline
    \textbf{\# of Prompts} & \textbf{Dice} & \textbf{TRE} \\ 
    \hline
    1 & 70.07±3.61 & 4.07±2.03 \\
    \rowcolor[rgb]{0.753,0.753,0.753}2 & 72.87±2.44 & 3.42±1.38 \\
    3 & 75.32±2.31 & 2.64±1.42 \\
    \rowcolor[rgb]{0.753,0.753,0.753}4 & 76.72±2.33 & 2.27±1.29 \\
    5 & 77.34±2.37 & 2.06±1.25 \\
    \rowcolor[rgb]{0.753,0.753,0.753}6 & 77.84±2.32 & 2.00±1.21 \\
    \hline
    \end{tabular}}
    \caption{Ablation of given prompt $Z^x_k$ quantity on MR-Prostate dataset.}
    \label{tab:x_number}
\end{minipage}
\hfill
\begin{minipage}{0.55\textwidth}  
    \centering
    \includegraphics[width=\textwidth]{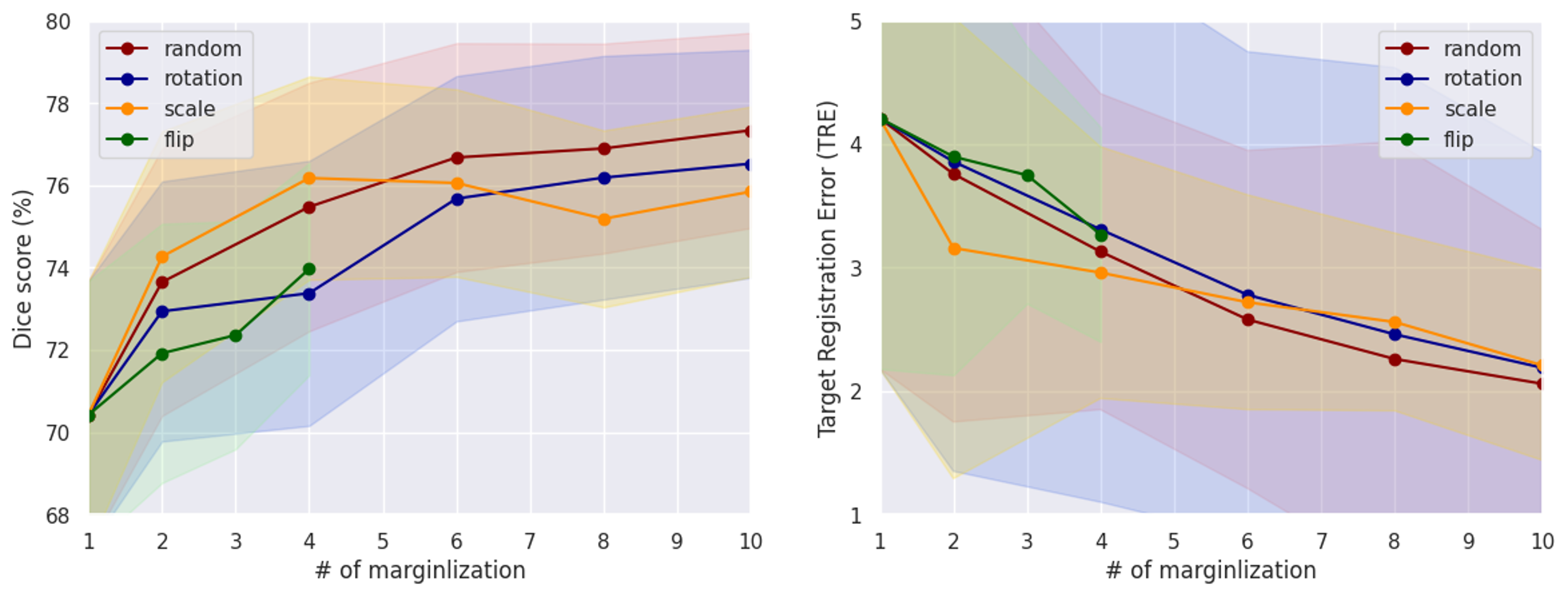}  
    \caption{Statistics of the number and type of marginalization iterations on MR-Prostate dataset.}
    \label{fig:margin_quant}
\end{minipage}
\vspace{-3em}
\end{table}

\noindent\textbf{Comparison with SoTA Methods.}
Table~\ref{tab:sota} presents a quantitative comparison across various medical image registration methods, including NiftyReg~\cite{modat2014global} that represents a class of non-learning iterative registration algorithms~\cite{796284}, SAMReg~\cite{huang2024one}, an existing SAM-based training-free algorithm, and also contrast them, albeit less equitably, unsupervised methods VoxelMorph, KeyMorph and lastest TransMorph~\cite{balakrishnan2019voxelmorph}, and ROI-supervised LabelReg~\cite{hu2018weakly}, on five datasets. 
Morph variants and LabelReg have been proposed for medical image research datasets, with LabelReg additionally requiring fully-segmented anatomical annotations. As shown in Table~\ref{tab:sota}, PromptReg consistently outperforms these methods across all datasets without dataset-specific tuning, particularly competitive in intra-subject registration tasks perhaps due to higher prevalence in consistent intra-subject ROIs.
Compared with SAMReg (5 paired ROIs), PromptReg demonstrates lower variance that may indicate greater robustness. Additionally, when local and/or dense correspondence is required clinically, the PromptReg exhibited superior capability to generate more paired ROIs that are indeed corresponding to each other - a limitation of SAMReg discussed in Sec.~\ref{sec: problem_def}.

\noindent\textbf{Comparison of Promptable Models.}
Promptable models can be directly integrated into the PromptReg paradigm, facilitating the adoption of cutting-edge segmentation models for registration. Table~\ref{tab:sota} evaluates various models with four random prompts each. For 3D images processed with 2D models (SAM, MedSAM, SAMed2d), slices are handled individually. Results show that 3D-specific models (SAMed3d, 3DAdapter) generally perform better due to their 3D awareness. SAM, however, excels in non-medical tasks due to its generalizability.
MedSAM and SAMed2d perform well on the Abdomen dataset, benefiting from pre-training on abdominal data. However, prior exposure doesn’t always improve registration, \textit{e.g.}, SAMed2d, pre-trained on prostate data, tends to focus on central structures, leading to inconsistencies when prompted outside these ROIs.

\noindent\textbf{Ablation on Given Prompt $Z^x_k$.}
\textbf{a}) Location. PromptReg allows predefined prompt locations. 
Constraining prompts within ROIs significantly improves registration, especially in inter-subject cases (Table~\ref{tab:x_loc}).
\textbf{b}) Quantity. More prompts enhance performance but with diminishing returns, improving ROI alignment and downstream tasks (Table~\ref{tab:x_number}).

\noindent\textbf{Ablation on Inverse Prompt $Z^y_k$.} 
\textbf{a}) Threshold $\sigma$. Table~\ref{tab:y_form} shows that $\sigma=20.0$ optimizes registration by balancing mismatch tolerance and auxiliary prompts.
\textbf{b}) Transformation $\mathcal{A}$. Fig.~\ref{fig:margin_quant} shows that scaling outperforms rotation for $<7$ iterations, while a random strategy works best for more. 

\section{Conclusion}
In this study, we introduce a new ``corresponding prompt problem'' that re-formulates the image registration to search corresponding prompts for pre-trained segmentation models, with a proposed promptable registration algorithm, PromptReg; comprehensive experiments highlight its competitive performance, indicating a new direction for registration research and a new application for vision foundation models.

\begin{credits}
\subsubsection{\ackname} This work was supported by the International Alliance for Cancer Early Detection, an alliance between Cancer Research UK [C28070/A30912; C73666/A31378], Canary Center at Stanford University, the University of Cambridge, OHSU Knight Cancer Institute, University College London and the University of Manchester. This work was also supported by the National Institute for Health Research (NIHR) University College London Hospitals (UCLH) Biomedical Research Centre (BRC).

\subsubsection{\discintname}
The authors have no competing interests to declare that are relevant to the content of this article.
\end{credits}
%
%
%
\bibliographystyle{splncs04}
\bibliography{Paper-4020}

\end{document}